\documentclass{article} 
\usepackage{nips14submit_e,times}
\usepackage{hyperref}
\usepackage{url}
\usepackage{graphicx}
\usepackage{wrapfig}
\usepackage{authblk}
\usepackage{listings}
\usepackage{booktabs}
\usepackage{amsmath}
\usepackage{amssymb}

\title{Simple Baseline for Visual Question Answering} 

\author[1]{Bolei Zhou}
\author[2]{Yuandong Tian}
\author[2]{Sainbayar Sukhbaatar}
\author[2]{Arthur Szlam}
\author[2]{Rob Fergus}
\affil[1]{Massachusetts Institute of Technology}
\affil[2]{Facebook AI Research}

\nipsfinalcopy 

\begin{document}
\maketitle

\begin{abstract}
We describe a very simple bag-of-words baseline for visual question answering. This baseline concatenates the word features from the question and CNN features from the image to predict the answer. When evaluated on the challenging VQA dataset~\cite{antol2015vqa}, it shows comparable performance to many recent approaches using recurrent neural networks. To explore the strength and weakness of the trained model, we also provide an interactive web demo\footnote{\url{http://visualqa.csail.mit.edu}}, and open-source  code\footnote{\url{https://github.com/metalbubble/VQAbaseline}}.
\end{abstract}

\section{Introduction}
Combining Natural Language Processing with Computer Vision for high-level scene interpretation is a recent trend, e.g., image captioning~\cite{mao2014deep,vinyals2014show,kiros2014multimodal,devlin2015exploring}. These works have
benefited from the rapid development of deep learning for visual recognition (object recognition \cite{krizhevsky2012imagenet} and scene recognition \cite{zhou2014learning}), and have been made possible by the emergence of large image datasets and text corpus (e.g., \cite{lin2014microsoft}). Beyond image captioning, a natural next step is visual question answering (QA) \cite{ren2015exploring,antol2015vqa,gao2015you}. 



Compared with the image captioning task, in which an algorithm is required to generate free-form text description for a given image, visual QA can involve a wider range of knowledge and reasoning skills. A captioning algorithm has the liberty to pick the easiest relevant descriptions of the image, whereas as responding to a question needs to find the correct answer for *that* question.   Furthermore, the algorithms for visual QA are required to answer all kinds of questions people might ask about the image, some of which might be relevant to the image contents, such as ``what books are under the television'' and ``what is the color of the boat'', while others might require knowledge or reasoning beyond the image content, such as ``why is the baby crying?'' and ``which chair is the most expensive?''.   Building robust algorithms for visual QA that perform at near human levels would be an important step towards solving AI.


Recently, several papers have appeared on arXiv (after CVPR'16 submission deadline) proposing neural network architectures for visual question answering, such as \cite{shih2015look,xu2015ask,gao2015you,yang2015stacked,wu2015ask,chen2015abc,noh2015image,andreas2015deep}. Some of them are derived from the image captioning framework, in which the output of a recurrent neural network (e.g., LSTM \cite{wu2015ask, noh2015image, andreas2015deep}) applied to the question sentence is concatenated with visual features from VGG or other CNNs to feed a classifier to predict the answer. Other models integrate visual attention mechanisms \cite{xu2015ask,shih2015look,chen2015abc} and visualize how the network learns to attend the local image regions relevant to the content of the question.   

Interestingly, we notice that in one of the earliest VQA papers \cite{ren2015exploring}, the simple baseline Bag-of-words + image feature (referred to as BOWIMG baseline) outperforms the LSTM-based models on a synthesized visual QA dataset built up on top of the image captions of COCO dataset \cite{lin2014microsoft}. For the recent much larger COCO VQA dataset \cite{antol2015vqa}, the BOWIMG baseline performs worse than the LSTM-based models \cite{antol2015vqa}. 

In this work, we carefully implement the BOWIMG baseline model. We call it iBOWIMG to avoid confusion with the implementation in \cite{antol2015vqa}. With proper setup and training, this simple baseline model shows comparable performance to many recent recurrent network-based approaches for visual QA. Further analysis shows that the baseline learns to correlate the informative words in the question sentence and visual concepts in the image with the answer. Furthermore, such correlations can be used to compute reasonable spatial attention map with the help of the CAM technique proposed in \cite{zhou2014learning}. The source code and the visual QA demo based on the trained model are publicly available. In the demo, iBOWIMG baseline gives answers to any question relevant to the given images. Playing with the visual QA models interactively could reveal the strengths and weakness of the trained model. 

\section{iBOWIMG for Visual Question Answering}
In most of the recent proposed models, visual QA is simplified to a classification task: the number of the different answers in the training set is the number of the final classes the models need to learn to predict. The general pipeline of those models is that the word feature extracted from the question sentence is concatenated with the visual feature extracted from the image, then they are fed into a softmax layer to predict the answer class. The visual feature is usually taken from the top of the VGG network or GoogLeNet, while the word features of the question sentence are usually the popular LSTM-based features \cite{ren2015exploring,antol2015vqa}. 

\begin{figure}
\begin{center}
\includegraphics[width=0.9\linewidth]{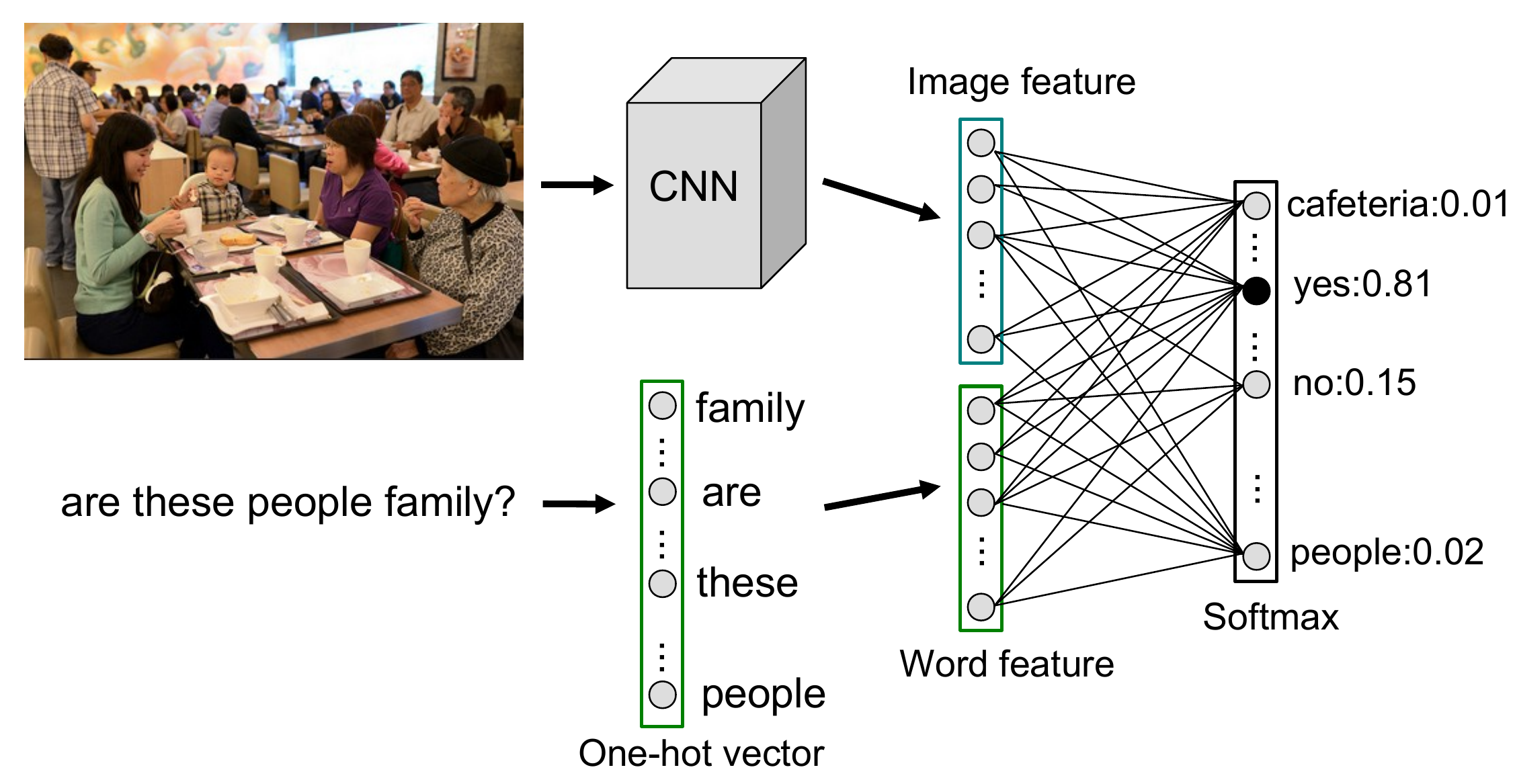}
\end{center}
\caption{Framework of the iBOWIMG. Features from the question sentence and image are concatenated then feed into softmax to predict the answer.}\label{vqa_framework}
\end{figure}

In our iBOWIMG model, we simply use naive bag-of-words as the text feature, and use the deep features from GoogLeNet \cite{szegedy2014going} as the visual features. Figure~\ref{vqa_framework} shows the framework of the iBOWIMG model, which can be implemented in Torch with no more than 10 lines of code. The input question is first converted to a one-hot vector, which is transformed to word feature via a word embedding layer and then is concatenated with the image feature from CNN. The combined feature is sent to the softmax layer to predict the answer class, which essentially is a multi-class logistic regression model.

\section{Experiments}
Here we train and evaluate the iBOWIMG model on the Full release of COCO VQA dataset \cite{antol2015vqa}, the largest VQA dataset so far. In the COCO VQA dataset, there are 3 questions annotated by Amazon Mechanical Turk (AMT) workers for each image in the COCO dataset. For each question, 10 answers are annotated by another batch of AMT workers. To pre-process the annotation for training, we perform majority voting on the 10 ground-truth answers to get the most certain answer for each question. Here the answer could be in single word or multiple words. Then we have the 3 question-answer pairs from each image for training. There are in total 248,349 pairs in train2014 and 121,512 pairs in val2014, for 123,287 images overall in the training set. Here train2014 and val2014 are the standard splits of the image set in the COCO dataset. 

To generate the training set and validation set for our model, we first randomly split the images of COCO val2014 into 70\% subset A and 30\% subset B. To avoid potential overfitting, questions sharing the same image will be placed into the same split. The question-answer pairs from the images of COCO train2014 + val2014 subset A are combined and used for training, while the val2014 subset B is used as validation set for parameter tuning. After we find the best model parameters, we combine the whole train2014 and val2014 to train the final model. We submit the prediction result given by the final model on the testing set (COCO test2015) to the evaluation server, to get the final accuracy on the test-dev and test-standard set. For Open-Ended Question track, we take the top-1 predicted answer from the softmax output. For the Multiple-Choice Question track, we first get the softmax probability for each of the given choices then select the most confident one.

The code is implemented in Torch. The training takes about 10 hours on a single GPU NVIDIA Titan Black.



\subsection{Benchmark Performance}

According to the evaluation standard of the VQA dataset, the result of the any proposed VQA models should report accuracy on the test-standard set for fair comparison. We report our baseline on the test-dev set in Table \ref{test-dev} and the test-standard set in Table \ref{test-standard}. The test-dev set is used for debugging and validation experiments and allows for unlimited submission to the evaluation server, while test-standard is used for model comparison with limited submission times.

Since this VQA dataset is rather new, the publicly available models evaluated on the dataset are all from non-peer reviewed arXiv papers. We include the performance of the models available at the time of writing (Dec.5, 2015) \cite{antol2015vqa, jiang2015compositional, andreas2015deep, shih2015look,wu2015ask,noh2015image}. Note that some models are evaluated on either test-dev or test-standard for either Open-Ended or Multiple-Choice track. 

The full set of the VQA dataset was released on Oct.6 2015;  previously the v0.1 version and v0.9 version had been released. We notice that some models are evaluated using non-standard setups, rendering performance comparisons difficult. \cite{xu2015ask} (arXiv dated at Nov.17 2015) used v0.9 version of VQA with their own split of training and testing; \cite{yang2015stacked} (arXiv dated at Nov.7 2015) used their own split of training and testing for the val2014; \cite{chen2015abc} (arXiv dated at Nov.18 2015) used v0.9 version of VQA dataset. So these are not included in the comparison.

\begin{table}
\small
\begin{center}
\caption{Performance comparison on test-dev.}\label{test-dev}
    \begin{tabular}{c c c c c}
    \toprule
    & \multicolumn{4}{ c }{Open-Ended } \\
    \hline
     &  \textbf{Overall} & yes/no & number & others \\
    \hline
    IMG \cite{antol2015vqa} & 28.13 & 64.01 & 00.42 & 03.77\\
    BOW \cite{antol2015vqa} & 48.09 & 75.66 & 36.70 & 27.14\\
    BOWIMG \cite{antol2015vqa} & 52.64 & 75.55 & 33.67 & 37.37\\
    LSTMIMG \cite{antol2015vqa}  & 53.74 & 78.94 & 35.24 & 36.42 \\
    CompMem \cite{jiang2015compositional} & 52.62 & 78.33 & 35.93 & 34.46 \\
    NMN+LSTM \cite{andreas2015deep} & 54.80 & 77.70 & 37.20 & 39.30 \\
    WR Sel. \cite{shih2015look} & - & - & - & - \\
    ACK \cite{wu2015ask} & 55.72 & 79.23 & 36.13 & 40.08  \\
	DPPnet \cite{noh2015image}   & \textbf{57.22} & 80.71 & 37.24 & 41.69 \\
    \hline
    iBOWIMG &  55.72   & 76.55 & 35.03 & 42.62 \\
    \bottomrule
    \end{tabular}
    \begin{tabular}{c c c c}
    \toprule
    \multicolumn{4}{ c }{Multiple-Choice} \\
    \hline
     \textbf{Overall} & yes/no & number & others \\
    \hline
	30.53 & 69.87 & 00.45 & 03.76 \\
    53.68 & 75.71 & 37.05 & 38.64 \\
    58.97 & 75.59 & 34.35 & 50.33 \\
    57.17 & 78.95 & 35.80 & 43.41\\
    - & - & - & - \\
    - & - & - & - \\
    60.96 & - & - & -\\
    - & - & - & - \\
    \textbf{62.48} & 80.79 & 38.94 & 52.16 \\
    \hline
    61.68 & 76.68 & 37.05 & 54.44\\
    \bottomrule
    \end{tabular}
    \end{center}
\end{table}

\begin{table}
\small
\begin{center}
\caption{Performance comparison on test-standard.}\label{test-standard}
    \begin{tabular}{c c c c c}
    \toprule
    & \multicolumn{4}{ c }{Open-Ended} \\
    \hline
     &  \textbf{Overall} & yes/no & number & others \\
    \hline
    LSTMIMG \cite{antol2015vqa} & 54.06 & - &- &- \\
	NMN+LSTM \cite{andreas2015deep}    & 55.10 & - &- &- \\
	ACK \cite{wu2015ask} &  55.98 & 79.05 & 36.10 & 40.61 \\
    DPPnet \cite{noh2015image} & \textbf{57.36} & 80.28 & 36.92 & 42.24 \\
    \hline
    iBOWIMG &  55.89   & 76.76 & 34.98 & 42.62 \\
    \bottomrule
    \end{tabular}
    \begin{tabular}{c c c c}
    \toprule
    \multicolumn{4}{ c }{Multiple-Choice} \\
    \hline
     \textbf{Overall}  & yes/no & number & others \\
    \hline
     -   & - & - & - \\
     -   & - & - & - \\
      -   & - & - & - \\
    \textbf{62.69} & 80.35 & 38.79 & 52.79 \\
    \hline
     61.97 & 76.86 & 37.30 & 54.60 \\
    \bottomrule
    \end{tabular}
    \end{center}
\end{table}

Except for these IMG, BOW, BOWIMG baselines provided in the \cite{antol2015vqa}, all the compared methods use either deep or recursive neural networks. However, our iBOWIMG baseline shows comparable performances against these much more complex models, except for DPPnet \cite{noh2015image} that is about 1.5\% better.


\subsection{Training Details}
\textbf{Learning rate and weight clip.} We find that setting up a different learning rate and weight clipping for the word embedding layer and softmax layer leads to better performance. The learning rate for the word embedding layer should be much higher than the learning rate of softmax layer to learn a good word embedding. From the performance of BOW in Table \ref{test-dev}, we can see that a good word model is crucial to the accuracy, as BOW model alone could achieve closely to 48\%, even without looking at the image content. 

\textbf{Model parameters to tune.} Though our model could be considered as the simplest baseline so far for visual QA, there are several model parameters to tune: \textbf{1)} the number of epochs to train. \textbf{2)} the learning rate and weight clip. \textbf{3)} the threshold for removing less frequent question word and answer classes. We iterate to search the best value of each model parameter separately on the val2014 subset B. In our best model, there are 5,746 words in the dictionary of question sentence, 5,216 classes of answers. The specific model parameters can be found in the source code.


\subsection{Understanding the Visual QA model}

From the comparisons above, we can see that our baseline model performs as well as the recurrent neural network models on the VQA dataset. Furthermore, due to its simplicity, the behavior of the model could be easily interpreted, demonstrating what it learned for visual QA. 

Essentially, the BOWIMG baseline model learns to memorize the correlation between the answer class and the informative words in the question sentence along with the visual feature. We split the learned weights of softmax into two parts, one part for the word feature and the other part for the visual feature. Therefore, 
\begin{align}\label{eq:linearscore}
r = \textbf{M}_{w}\textbf{x}_{w} + \textbf{M}_{v}\textbf{x}_{v}.
\end{align}
Here the softmax matrix $\textbf{M}$ is decomposed into the weights $\textbf{M}_w$ for word feature $\textbf{x}_{w}$ and the weights $\textbf{M}_v$ for the visual feature $\textbf{x}_{v}$ whereas $\textbf{M} = [\textbf{M}_w,\textbf{M}_v]$. $r$ is the response of the answer class before softmax normalization. Denote the response $r_w = \textbf{M}_{w}\textbf{x}_{w}$ as the contribution from question words and $r_v = \textbf{M}_{v}\textbf{x}_{v}$ as the contribution from the image contents. Thus for each predicted answer, we know exactly the proportions of contribution from word and image content respectively. We also could rank $r_w$ and $r_v$ to know what the predicted answer could be if the model only relies on one side of information. 

\begin{figure}
\begin{center}
\includegraphics[width=1\linewidth]{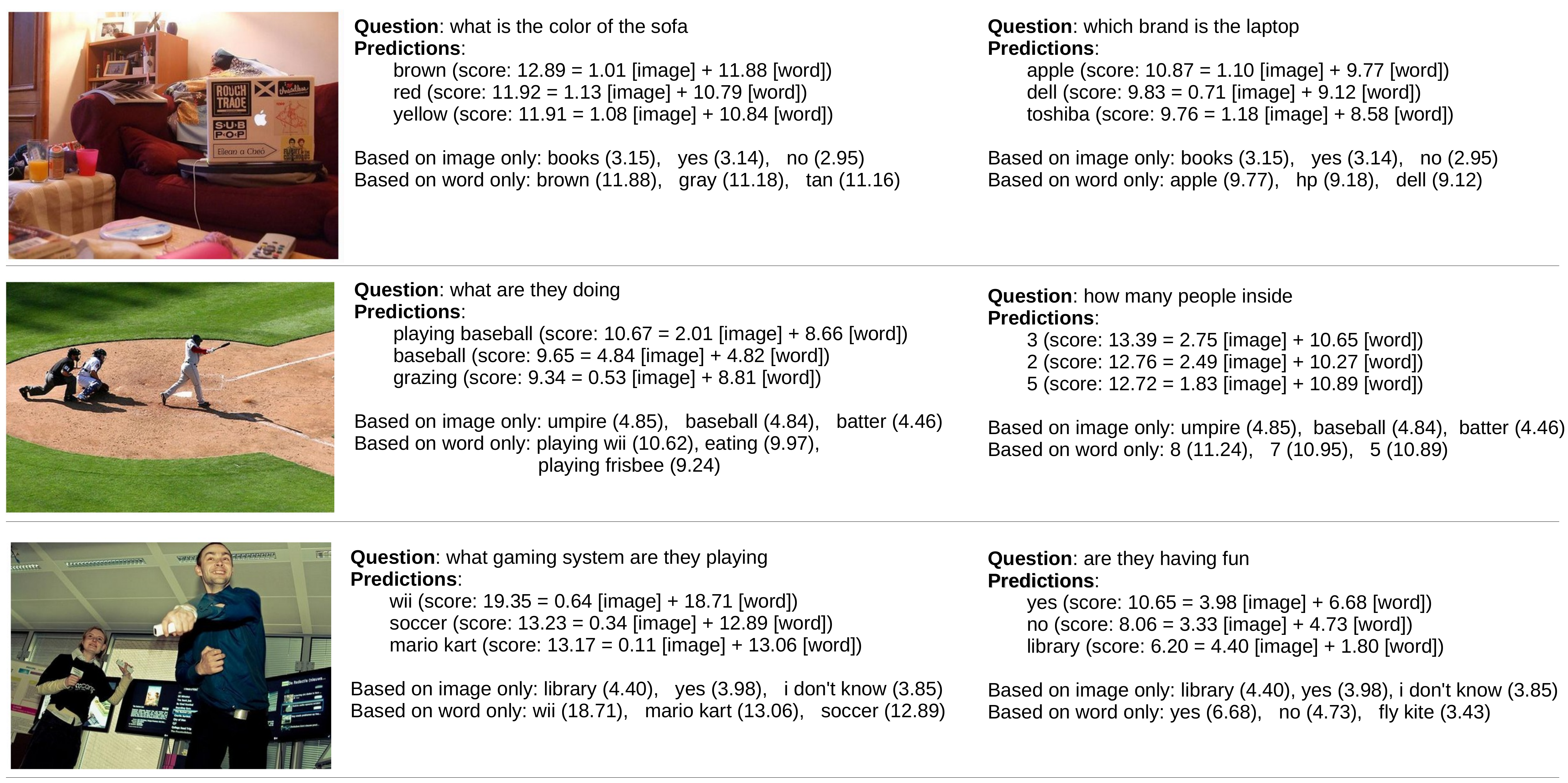}
\end{center}
\caption{Examples of visual question answering from the iBOWIMG baseline. For each image there are two questions and the top 3 predicted answers from the model. The prediction score of each answer is decomposed into the contributions of image and words respectively. The predicted answers which rely purely on question words or image are also shown.}\label{vqa_prediction}
\end{figure}

Figure~\ref{vqa_prediction} shows some examples of the predictions, revealing that the question words usually have dominant influence on predicting the answer. For example, the correctly predicted answers for the two questions given for the first image `what is the color of sofa' and `which brand is the laptop' come mostly from the question words, without the need for image. This demonstrates the bias in the frequency of object and actions appearing in the images of COCO dataset. For the second image, we ask `what are they doing': the words-only prediction gives `playing wii (10.62), eating (9.97), playing frisbee (9.24)', while full prediction gives the correct prediction `playing baseball (10.67 = 2.01 [image] + 8.66 [word])'.  

To further understand the answers predicted by the model given the visual feature and question sentence, we first decompose the word contribution of the answer into single words of the question sentence, then we visualize the informative image regions relevant to the answer through the technique proposed in \cite{zhou2015localizable}. 

\begin{figure}
\begin{center}
\includegraphics[width=0.9\linewidth]{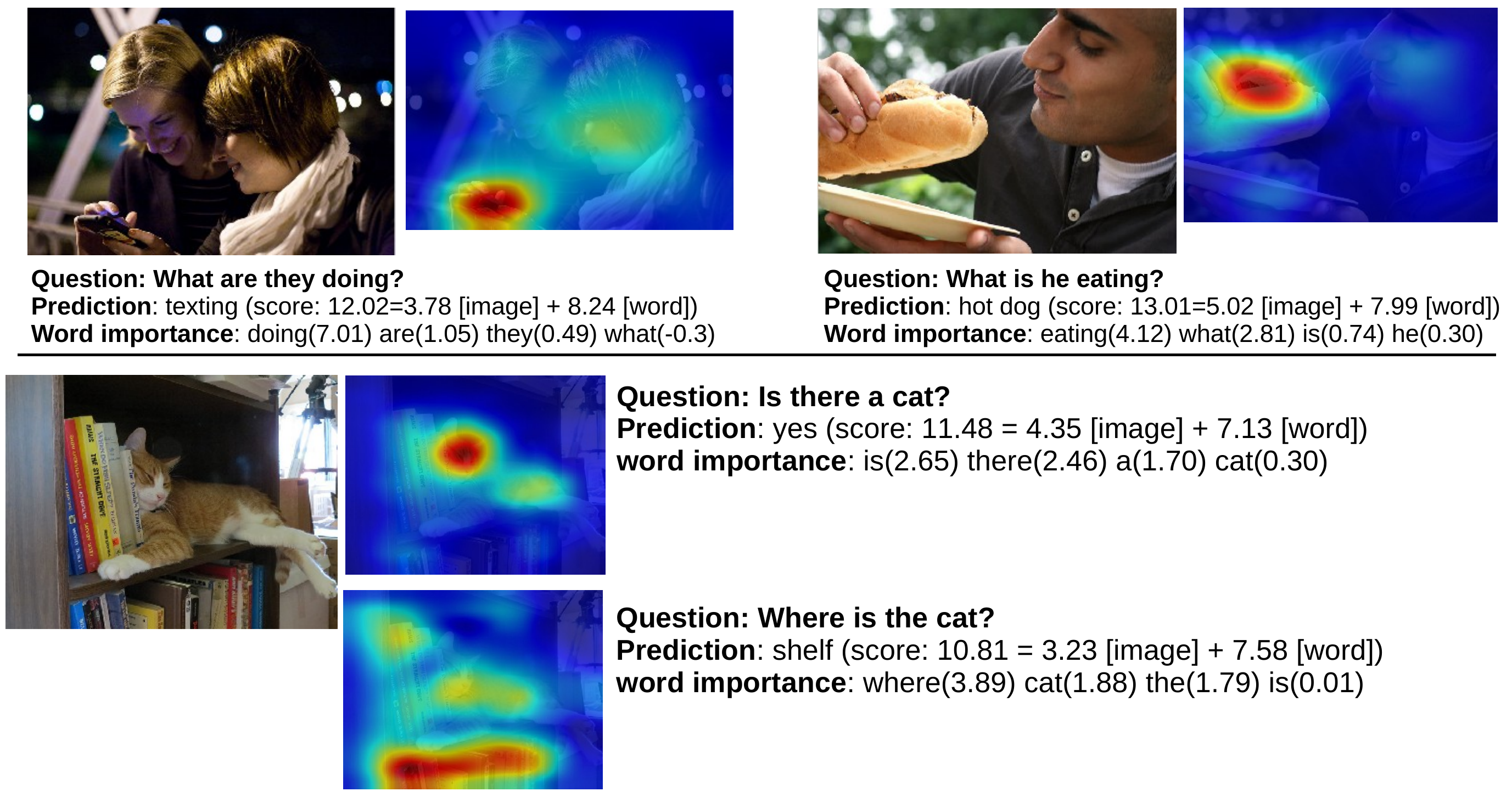}
\end{center}
\caption{The examples of the word importance of question sentences and the informative image regions relevant to the predicted answers.}\label{CAM}
\end{figure}

Since there are just two linear transformations (one is word embedding and the other is softmax matrix multiplication) from the one hot vector to the answer response, we could easily know the importance of each single word in the question to the predicted answer. In Figure~\ref{CAM}, we plot the ranked word importance for each word in the question sentence. In the first image question word `doing' is informative to the answer `texting' while in the second image question word `eating' is informative to the answer `hot dog'. 

To highlight the informative image regions relevant to the predicted answer we apply a technique called Class Activation Mapping (CAM) proposed in \cite{zhou2015localizable}. The CAM technique leverages the linear relation between the softmax prediction and the final convolutional feature map, which allows us to identify the most discriminative image regions relevant to the predicted result. In Figure~\ref{CAM} we plot the heatmaps generated by the CAM associated with the predicted answer, which highlight the informative image regions such as the cellphone in the first image to the answer `texting' and the hot dog in the first image to the answer `hot dog'. The example in lower part of Figure~\ref{CAM} shows the heatmaps generated by two different questions and answers. Visual features from CNN already have implicit attention and selectivity over the image region, thus the resulting class activation maps are similar to the maps generated by the attention mechanisms of the VQA models in \cite{shih2015look,xu2015ask,yang2015stacked}. 

\section{Interactive Visual QA Demo} 

\begin{figure}
\begin{center}
\includegraphics[width=1\linewidth]{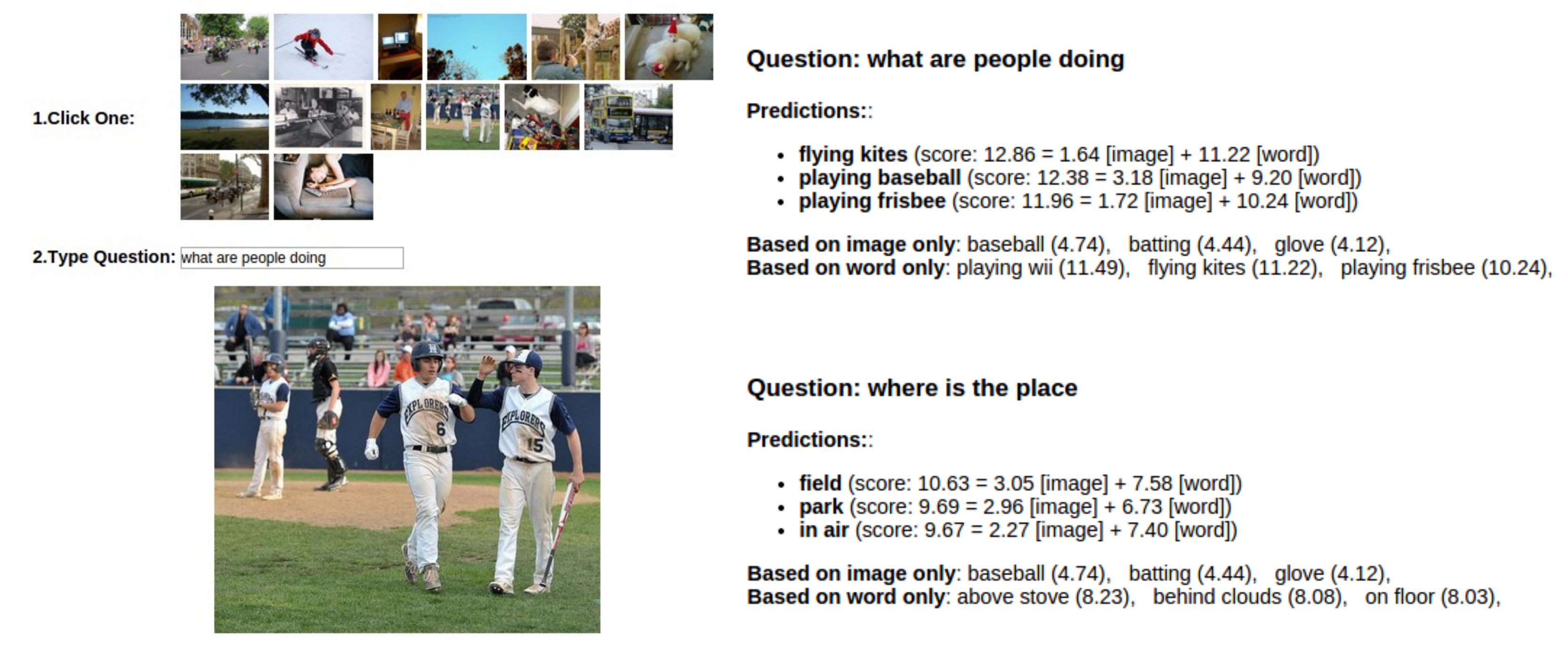}
\end{center}
\caption{Snapshot of the visual question answering demo. People could type questions into the demo and the demo will give answer predictions. Here we show the answer predictions for two questions.}\label{vqa_demo}
\end{figure}

Question answering is essentially an interactive activity, thus it would be good to make the trained models able to interact with people in real time. Aided by the simplicity of the baseline model, we built a web demo that people could type question about a given image and our AI system powered by iBOWIMG will reply the most possible answers. Here the deep feature of the images are extracted beforehand. Figure~\ref{vqa_demo} shows a snapshot of the demo. People could play with the demo to see the strength and weakness of VQA model.

\section{Concluding Remarks}


For visual question answering on COCO dataset, our implementation of a simple baseline achieves comparable performance to several recently proposed recurrent neural network-based approaches. To reach the correct prediction, the baseline captures the correlation between the informative words in the question and the answer, and that between image contents and the answer. How to move beyond this, from memorizing the correlations to actual reasoning and understanding of the question and image, is a goal for future research.

\bibliographystyle{ieee}
\bibliography{reference}
\end{document}